\documentclass[10pt,twocolumn,letterpaper]{article}

\usepackage{multirow}
\usepackage{cvpr}
\usepackage{times}
\usepackage{epsfig}
\usepackage{graphicx}
\usepackage{amsmath}
\usepackage{amssymb}
\usepackage[export]{adjustbox}
\usepackage[numbers,sort]{natbib} 

\usepackage{multido}
\usepackage{subcaption}
\usepackage{comment}

\usepackage[pagebackref=true,breaklinks=true,letterpaper=true,colorlinks,bookmarks=false]{hyperref}

\cvprfinalcopy

\ifcvprfinal\fi
\begin{document}

\title{A Short Note on the  Kinetics-700 Human Action Dataset}

\author{{Jo\~a}o Carreira \\
{\tt\small joaoluis@google.com} \\
\and
Eric Noland \\
{\tt\small enoland@google.com} \\
\and
Chloe Hillier \\
{\tt\small chillier@google.com} \\
\and
Andrew Zisserman \\
{\tt\small zisserman@google.com} \\
}

\maketitle

\begin{abstract}

We describe an extension of the DeepMind Kinetics human action dataset
from 600 classes to 700 classes, where for each class there are at least 600 video clips from different
YouTube videos.
This paper details the changes introduced for this new release of the dataset, 
and includes a comprehensive set of statistics as well as baseline results using the I3D neural network
architecture. 
\end{abstract}

\section{Introduction}

The goal of the Kinetics project is to provide a large scale curated
dataset of video clips, covering a diverse range of human actions,
that can be used for training and exploring neural network
architectures for modelling human actions in video.  This short paper
describes the new version of the dataset, called Kinetics-700.

The new dataset follows the same
principles as Kinetics-400~\cite{kay2017kinetics} and Kinetics-600~\cite{kinetics600}: (i) The clips are from
YouTube videos, last 10s, and have a variable resolution and frame
rate; (ii) for an action class, all clips are from different YouTube
videos.  Kinetics-700 is almost a superset of Kinetics-600: the number of classes is increased from 600 to 700, with all but
three of the Kinetics-600 classes retained. As in the case of Kinetics-600,  Kinetics-700 has
600 or more clips per human action class -- this 
represents a  30\% increase in the number of video clips,
from around 500k to around 650k. The statistics of the three Kinetics datasets
are detailed in table~\ref{tab:kinetics_stats}.

In the new Kinetics-700 dataset there is a standard validation set,
for which labels have been publicly released, and also a held-out test
set (where the labels are not released).  We encourage researchers to
report results on the standard validation set, unless they want to
compare with participants of the Activity-Net Kinetics
challenge where the performance on the held-out test set can be be measured only
through the challenge evaluation
website\footnote{\url{http://activity-net.org/challenges/2019/evaluation.html}}. 
The URLs of Kinetics YouTube videos and temporal intervals can be obtained from \url{http://deepmind.com/kinetics}.

\begin{table*}[ht]
\centering
\begin{tabular}{| l || c | c | c | c || c | c | c | }
  \hline
  \textbf{Version} & \textbf{Train} & \textbf{Valid.} & \textbf{Test} & \textbf{Held-out Test} & \textbf{Total Train} & \textbf{Total} & {\textbf{Classes}}  \\ \hline 
Kinetics-400 \cite{kay2017kinetics} & 250--1000 & 50  & 100 & 0  & 246,245 & 306,245 & 400 \\ \hline
Kinetics-600 \cite{kinetics600} & 450--1000 & 50  & 100 & around 50 & 392,622 & 495,547 & 600 \\  \hline
Kinetics-700 & 450--1000 & 50 & 100 & 0  &  545,317   &  650,317  & 700 \\  \hline
\end{tabular} 
\vspace{5pt}
\caption{Kinetics Dataset Statistics. The number of clips for each class in the various splits (left), and the totals (right). }
\label{tab:kinetics_stats}
\end{table*}

\section{Data Collection Process \label{collection}}

The data collection process evolved from Kinetics-400 to
Kinetics-700, although the overall pipeline is  the same: 1) action class
sourcing, 2) candidate video matching, 3) candidate clip selection, 4)
human verification, 5) quality analysis and filtering. In words, a
list of class names is created, then a list of candidate YouTube URLs
is obtained for each class name, and candidate 10s clips are sampled
from the videos. These clips are sent to humans who
decide whether those clips contain the action class that they are
supposed to. Finally, there is an overall curation process including
clip de-duplication, and selecting the higher quality classes and
clips. Full details can be found in the original
publication~\cite{kay2017kinetics}.

The main differences in the data collection process between
Kinetics-400, Kinetics-600 and 700 is  in steps 1, 2 and 4: how action classes
are sourced, how candidate YouTube videos are matched with classes,  and human verification. In the following we detail these differences and the consequences of these changes on  the dataset.
Note, as well as producing clips for entirely new classes, it is necessary to `top up' existing classes in Kinetics-600 since
YouTube videos are deleted or unlisted over time (about 3\% per year).

It should be noted that the design of the collection process is not
well suited to finding action classes that progress over time. It is
very well suited to continual actions that exist over the length of the video
(e.g.\ `juggling', `drumming'), but not to those that have a
progression from start to middle to end (e.g.\ `dropping plates', `getting out of car').

\vspace{3mm}
\subsection{Action class sourcing} 
The additional classes for Kinetics-700 over Kinetics-600 were partly
sourced from the lists of actions (or verbs) in recent human action
datasets, such as EPIC-Kitchens~\cite{Damen_2018_ECCV} and AVA~\cite{gu2017ava}. Also, some
existing classes in Kinetics-600 which were at quite a general level,
e.g.\  `picking fruit', were removed and
replaced by a number of fine-grained variations, for example: `picking apples',
`picking blueberries'.
We also introduced a number of more imaginative classes, such as:
`making slime', `being in zero gravity', `swimming with sharks'.

\vspace{3mm}
\subsection{Candidate video matching} 
In Kinetics-700 we formally separated the `class name'  from  the `query text' used to search for that class.
So, for example, to obtain the class `canoeing or kayaking', the query text could be canoeing and kayaking, and both would be used. Another example is `abseiling', which can be queried with both `abseiling' or `rappelling'. Further more,
the query text was translated into three languages. In Kinetics-600 we had piloted this scheme by using both
English and Portuguese query texts, but in Kinetics-700 we extended it.  We describe next these multiple
queries and how they are matched to the YouTube corpus to obtain candidate videos.

\vspace{3mm}
\noindent \textbf{Multiple queries.} 
In order to get a better and larger pool of candidates for a class, each query text was automatically
translated from English into three languages: French, Portuguese, and Spanish. 
These are three out of
six languages with the most native speakers in the
world\footnote{According to
https://www.babbel.com/en/magazine/the-10-most-spoken-languages-in-the-world/},
and have large YouTube communities. We found that the machine translation had adequate quality, though sometimes
it introduced ambiguity. The query texts in all four languages were used to obtain candidate videos.

Having multiple languages had the positive side effect of also
promoting slightly greater dataset diversity by incorporating a more
well-rounded range of cultures, ethnicities and geographies. In terms of continents, more than 50\% of the clips are sourced from North America. However,  the fraction of clips from Latin America increased from 3\% in Kinetics-400 to 8\% in Kinetics-700, thanks to adding Spanish and Portuguese language queries. Africa is still the least represented continent, increasing from 0.8\% in Kinetics-400 to 1\% in Kinetics-700. These numbers are based on the 90\% of videos that contained location information.

\vspace{3mm}
\noindent \textbf{Matching query text to YouTube videos.} 
Rather than matching directly using textual queries we found it
beneficial to use weighted ngram representations of the combination of
the metadata of each video and the titles of related
ones. Importantly, these representations were compatible with multiple
languages. We combined this with standard title matching to get a
robust similarity score between a query and all YouTube videos.
This  meant that  we never ran out of
candidates, although the human-verification yield of the selected
candidates became lower for smaller similarity values. This procedure generates a far larger candidate
pool than simply a binary match between the query text and YouTube video title, say.
Since the target length of a clip is 10s, videos longer than 5 minutes were discouraged.

\subsection{Candidate clip selection and yield} 
Within a video,  candidate clips are selected by using image classifiers.
Image classifiers are available for a large number of human actions.
These classifiers are obtained by tracking user actions on Google
Image Search.  For example, for a search query ``climbing tree'', user
relevance feedback on images is collected by aggregating across the
multiple times that search query is issued. This relevance
feedback is used to select a high-confidence set of images that can
be used to train a ``climbing tree'' image classifier.  
Classifiers corresponding to the class name are run at the frame level over the selected videos for that class,
and clips extracted around the top $k$ responses (where $k=2$). In cases where we could not find classifiers for the class name, we used classifiers related to the query texts.

\subsection{Human verification}

The first and main annotation task in our pipeline asks human annotators if a clip contains a particular action. This step was the same as in previous years for Kinetics-700.

A difference to previous years was in the final human annotation
stage, which we previously did not crowdsource and instead did
ourselves: we would go over each individual class and look at all its
animated-gif thumbnails while taking into account potentially confusing classes
(derived from classifier outputs). Sometimes class names may allow for
multiple types of videos -- e.g.\  a class named ``jumping into pool"
could have people diving or just jumping. If we had a competing
``diving" class then we would try to remove diving videos from ``jumping
into pool".

This was a painstaking manual effort, which we tried to crowdsource this year. Since crowdsourcing requires limiting the size of individual tasks, we divided class thumbnails into panels of 16 elements and had human workers clean up the classes. Note that this provides them however with a tighter window into the contents of each class.

\vspace{3mm}
\noindent \textbf{Yield by class.} It is interesting to see which classes gave the highest and lowest yields in terms
of the probability that a candidate clip was voted positive for that class by three or more human annotators.
The classes with highest yield are given in table~\ref{tab:highyield}, and those with lowest yield are listed
in Appendix~\ref{sec:lowyield}.

\begin{table}
\centering
\begin{tabular}{|r| l| c | }
  \hline
  \textbf{Rank } &   \textbf{Class } & \textbf{Yield}  \\ \hline
1~~ &   busking	& 0.9227	\\ 
2~~ &  spinning poi	& 0.9227	\\
3~~ &  rope pushdown	& 0.9091	\\
4~~ &   front raises	& 0.8864	\\
5~~ &   zumba	& 0.8864	\\
6~~ &   country line dancing	& 0.8727	\\
7~~ &   ice skating	& 0.8636	\\
8~~ &   shearing sheep	& 0.8636	\\
9~~ &   arm wrestling	& 0.8636	\\
10~~ &   bench pressing	& 0.8545	\\
11~~ &   playing squash or racquetball	& 0.8455	\\
12~~ &   playing accordion	& 0.8318	 \\
\hline
\end{tabular} 
\vspace{5pt}
\caption{The classes that have the highest yield -- measured as the proportion of candidate clips that were judged positive
for that class by three or more annotators.}
\label{tab:highyield}
\end{table}

There are multiple factors involved here: whether the query is text that is used to annotate videos;
how general or specific the query text is for obtaining relevant videos
(for example, ``acting in play" is already quite ambiguous in English, and the current automatic translations are totally off, e.g.\ for Portuguese it translates into what would translatee back as ``acting in game"); how well the clip is selected within a relevant video; and the actual numbers of videos on YouTube for
that action class. One notable common element of the highest yields is that they are the type
of actions where the temporal position selected is not important -- `playing guitar' will be true at almost any point over
a long temporal period, and the video is easily specified by the class name; in contrast `opening a letter' only occurs
over a very specific and short time interval, and consequently could easily be missed in a long video.

In general the high yield classes are successful in being included in
the Kinetics release, but conversely, only a small proportion of the
low yield classes survives.

\section{From Kinetics-600 to Kinetics-700}

As mentioned above,  Kinetics-700 is an approximate superset of Kinetics-600 -- overall,
597 out of 600 classes are exactly the same in
Kinetics-700 (although some of the clips may have been replaced if the original videos have been deleted). For the other
classes, we renamed one (``passing american football (not in game)" to ``passing American football (not in game)"), and 
split ``chopping vegetables" and 
``picking fruit" into multiple subclasses.

In terms of the train/val/test split, there is a very small overlap between the Kinetics-700 test set and Kinetics-600 train/val/test/hold out test (under 3\%).

It is therefore largely \textbf{safe} to use models that have been trained on Kinetics-600 to evaluate the Kinetics-700 test set (the activity-net evaluation website explicitly ignores the predictions on those 3\% clips when evaluating on the test set). It is however \textbf{not safe} to train on Kinetics-700 and then evaluate on Kinetics-600 -- many of the validation/test clips from Kinetics-600 are in the training set of Kinetics-700~\footnote{This is because we created a fresh held-out test set for Kinetics-700.}!

The full list of new classes in Kinetics-700 is given in  Appendix~\ref{sec:newclasses}.

\section{Benchmark Performance}

\begin{table}
\centering
\begin{tabular}{| c| r | r | c |}
  \hline
  \textbf{Acc. type} & \textbf{Valid} & \textbf{Test}  \\ \hline
Top-1 & 58.7 & 57.3  \\ \hline
Top-5 & 81.7 & 79.9  \\ \hline
$100.0-avg$(Top-1,Top-5) & 29.8 & 31.4 \\ \hline
\end{tabular} 
\vspace{5pt}
\caption{Performance of an I3D model with RGB inputs on the Kinetics-700 dataset, without any test time augmentation (processing a center crop of each video convolutionally in time). The first two rows show accuracy in percentage, the last one shows the metric used at the Kinetics challenge hosted by the ActivityNet workshop.}
\label{tab:benchmark}
\end{table}

\begin{figure}
  \centering
    \includegraphics[width=0.45\textwidth]{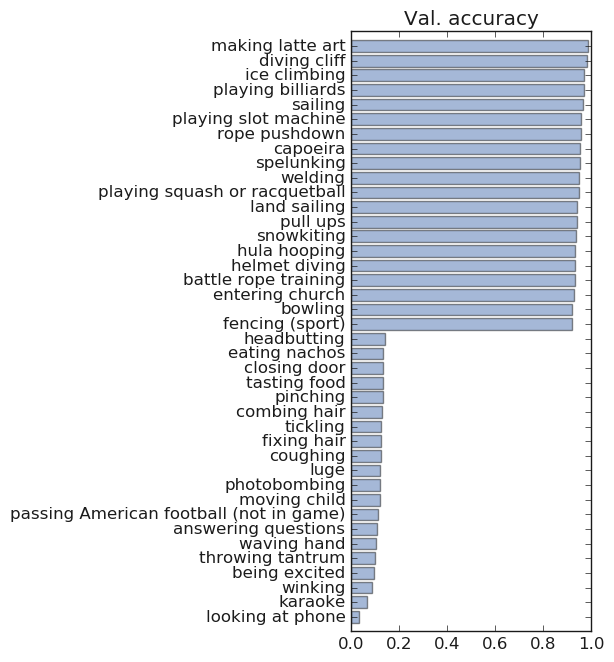}
    \caption{List of 20 easiest and 20 hardest Kinetics-700 classes sorted by class accuracies obtained using the I3D-RGB model.}
    \label{fig:accuracies}
\end{figure}

As a baseline model we used I3D \cite{Carreira17}, with standard RGB videos as input (no optical flow). 
We trained the model from scratch on the Kinetics-700 training set, picked hyper-parameters on validation, and report performance on validation and test set. 
We used 32 P100 GPUs, batch size 5 videos, 64 frame clips for training and 251 frames for testing. We trained using SGD with momentum, starting with a learning rate of 0.1, decreasing it by a factor of 10 when the loss saturates. Results are shown in table~\ref{tab:benchmark}. Hardest and easiest classes are shown in fig.~\ref{fig:accuracies}.

The top-1 accuracy on the validation set was 58.7 and on the test set was 57.3, which shows that both sets are similarly hard. On Kinetics-400 the corresponding test set accuracy was 68.4 and on Kinetics-600 it was 71.7, hence the task overall seems to have became considerably harder. This may partially have to do with the way we have now crowdsourced the human verification stage -- it may be that workers did not strive as hard as we previously did to make classes more unimodal. It is possible also that, since we collected a full new test set, there is a little distribution shift between train and test (but the validation performance is not very different from the test performance).
 
\vspace{3mm}
\noindent \textbf{Kinetics challenge.} 
There was a first Kinetics challenge at the ActivityNet workshop in
CVPR 2017, using Kinetics-400. The second challenge occurred at the
ActivityNet workshop in CVPR 2018, this time using Kinetics-600. The
performance criterion used in the challenge is the average of Top-1
and Top-5 error. There was an  improvement between the winning systems
of the two challenges, with error getting down from 12.4\% (in 2017) to 11.0\% 
(in 2018)~\cite{bian2017revisiting,he2018exploiting}. The 2019 challenge featured the new Kinetics-700 dataset and had 15 participating teams. The top team was from JD AI Research and obtained 17.9\% error, considerably below our baseline -- a single RGB I3D model -- which obtained 29.3\% error.

\section{Conclusion}
\label{sec:conclusion}

We have described the new Kinetics-700 dataset, which in terms of clip counts is 30\% larger
than Kinetics-600, and more than doubles the size of the original Kinetics-400 dataset.  It represents another step
towards our original goal of producing an action classification dataset with
1000 classes. Slides illustrating some of what has been described in this paper can be found online\footnote{\url{https://drive.google.com/file/d/164kU_MFTKzmefbgOLntuiiTmADutl_x0/view}}.

\subsection*{Acknowledgements:} The collection of this dataset was funded by DeepMind. 
The authors would like to thank Jean-Baptiste Alayrac for checking and correcting the French machine translated queries.

{\small
\bibliographystyle{ieee}
\bibliography{references}

\begin{thebibliography}{1}\itemsep=-1pt

\bibitem{bian2017revisiting}
Y.~Bian, C.~Gan, X.~Liu, F.~Li, X.~Long, Y.~Li, H.~Qi, J.~Zhou, S.~Wen, and
  Y.~Lin.
\newblock Revisiting the effectiveness of off-the-shelf temporal modeling
  approaches for large-scale video classification.
\newblock {\em arXiv preprint arXiv:1708.03805}, 2017.

\bibitem{kinetics600}
J.~Carreira, E.~Noland, A.~Banki-Horvath, C.~Hillier, and A.~Zisserman.
\newblock A short note about {Kinetics-600}.
\newblock {\em arXiv preprint arXiv:1808.01340}, 2018.

\bibitem{Carreira17}
J.~Carreira and A.~Zisserman.
\newblock {Quo Vadis, Action Recognition? New Models and the Kinetics Dataset}.
\newblock In {\em IEEE International Conference on Computer Vision and Pattern
  Recognition CVPR}, 2017.

\bibitem{Damen_2018_ECCV}
D.~Damen, H.~Doughty, G.~Maria~Farinella, S.~Fidler, A.~Furnari, E.~Kazakos,
  D.~Moltisanti, J.~Munro, T.~Perrett, W.~Price, and M.~Wray.
\newblock {Scaling Egocentric Vision: The EPIC-KITCHENS Dataset}.
\newblock In {\em Proc. ECCV}, 2018.

\bibitem{gu2017ava}
C.~Gu, C.~Sun, D.~A. Ross, C.~Vondrick, C.~Pantofaru, Y.~Li,
  S.~Vijayanarasimhan, G.~Toderici, S.~Ricco, R.~Sukthankar, et~al.
\newblock {AVA}: A video dataset of spatio-temporally localized atomic visual
  actions.
\newblock {\em CoRR, abs/1705.08421}, 4, 2017.

\bibitem{he2018exploiting}
D.~He, F.~Li, Q.~Zhao, X.~Long, Y.~Fu, and S.~Wen.
\newblock Exploiting spatial-temporal modelling and multi-modal fusion for
  human action recognition.
\newblock {\em arXiv preprint arXiv:1806.10319}, 2018.

\bibitem{kay2017kinetics}
W.~Kay, J.~Carreira, K.~Simonyan, B.~Zhang, C.~Hillier, S.~Vijayanarasimhan,
  F.~Viola, T.~Green, T.~Back, P.~Natsev, M.~Suleyman, and A.~Zisserman.
\newblock The kinetics human action video dataset.
\newblock {\em arXiv preprint arXiv:1705.06950}, 2017.

\end{thebibliography}
}

\appendix

\section{List of New Human Action Classes in Kinetics-700}
\label{sec:newclasses}
This is the list of classes in Kinetics-700 that were not in Kinetics-600,  or that have been renamed.

\begin{enumerate}
\itemsep0em 
\item pouring wine
\item walking on stilts
\item listening with headphones
\item dealing cards
\item sanding wood
\item splashing water
\item digging
\item chasing
\item tossing salad
\item playing cards
\item moving baby
\item bouncing ball (not juggling)
\item helmet diving
\item vacuuming car
\item high fiving
\item picking apples
\item swimming with sharks
\item cutting cake
\item doing sudoku
\item swimming with dolphins
\item playing american football
\item pouring milk
\item entering church
\item carrying weight
\item taking photo
\item saluting
\item jumping sofa
\item exercising arm
\item playing oboe
\item shooting off fireworks
\item playing nose flute
\item making latte art
\item carving wood with a knife
\item making slime
\item looking in mirror
\item shoot dance
\item checking watch
\item playing checkers
\item seasoning food
\item sieving
\item gargling
\item pulling espresso shot
\item curling eyelashes
\item shredding paper
\item stacking dice
\item surveying
\item poaching eggs
\item pulling rope (game)
\item uncorking champagne
\item eating nachos
\item picking blueberries
\item coughing
\item filling cake
\item shouting
\item playing mahjong
\item spinning plates
\item spraying
\item pretending to be a statue
\item moving child
\item steering car
\item baby waking up
\item treating wood
\item playing piccolo
\item letting go of balloon
\item playing shuffleboard
\item playing road hockey
\item using megaphone
\item squeezing orange
\item being in zero gravity
\item walking with crutches
\item polishing furniture
\item closing door
\item grooming cat
\item laying decking
\item arresting
\item rolling eyes
\item ski ballet
\item mixing colours
\item metal detecting
\item waxing armpits
\item peeling banana
\item cooking chicken
\item carving marble
\item filling eyebrows
\item breaking glass
\item playing rounders
\item petting horse
\item putting wallpaper on wall
\item herding cattle
\item playing billiards
\item stacking cups
\item blending fruit
\item lighting candle
\item decoupage
\item crocheting
\item playing slot machine
\item silent disco
\item being excited
\item brushing floor
\item opening coconuts
\item milking goat
\item slicing onion
\item flipping bottle

\end{enumerate}

\section{List of Low Yield Classes}
\label{sec:lowyield}
This is the ranked list of classes that have lowest yield, where yield is
the probability that a candidate clip was voted positive for that class by three or more human annotators.
Bold indicates that the class was included in the final dataset;  most of the low yield classes were not included.

\begin{enumerate}
\itemsep0em 
\item  opening letter	0.0019	
\item adding fish to aquarium	0.0033	
\item getting inside balloon	0.0034	
\item comforting	0.0036	
\item highlight text	0.0038	
\item riding giraffe	0.0047	
\item dropping plates	0.0057	
\item contemplating	0.0061	
\item whispering	0.0101	
\item grooming (person)	0.0106	
\item boarding train	0.0112	
\item buying fast food	0.0114	
\item Piling coins up	0.0114	
\item looking through telescope	0.0116	
\item breaking aquarium	0.0118	
\item using a crowbar	0.0127	
\item underlining	0.0128	
\item instant messaging	0.0133	
\item getting into a car	0.0134	
\item {\bf tossing coin}	0.0146
\item getting out of a car	0.0153	
\item checking mail	0.0157	
\item entering building	0.0177	
\item signing document	0.0179	
\item cutting in line	0.0179	
\item waiting at crossing	0.0179	
\item dunking biscuit	0.0185	
\item checking tickets	0.0188	
\item {\bf assembling bicycle}	0.0196
\item exiting building	0.0198	
\item unloading the trunk of a car	0.0198	
\item setting up fish tank	0.0198	
\item cutting squares	0.0201	
\item {\bf texting}	0.0209 
\item playing underwater frisbee	0.0212	
\item riding zebra	0.0212	
\end{enumerate}

\end{document}